\documentclass[11pt,a4paper]{article}
\usepackage[hyperref]{emnlp2018}
\usepackage{times}
\usepackage{latexsym}
\usepackage{color}
\usepackage{url}
\usepackage{amsmath}
\usepackage{amssymb}
\usepackage{booktabs}
\usepackage{multirow}
\usepackage{subcaption}
\usepackage{caption}
\usepackage{graphicx}
\usepackage{bigdelim}

\aclfinalcopy 
\def\aclpaperid{2043} 

\newcommand{\bd}[1]{\textcolor{blue}{Bhuwan: #1}} 

\title{Open Domain Question Answering Using\\Early Fusion of Knowledge Bases and Text}

\author{
Haitian Sun\thanks{\enskip Haitian Sun and Bhuwan Dhingra contributed equally to this work.}\qquad Bhuwan Dhingra\footnotemark[1]\qquad Manzil Zaheer\qquad Kathryn Mazaitis\\
{\bf Ruslan Salakhutdinov\qquad William W. Cohen}\\
School of Computer Science\\
Carnegie Mellon University\\
\texttt{\{haitians,bdhingra,manzilz,krivard,rsalakhu,wcohen\}@cs.cmu.edu}}

\date{}

\begin{document}
\maketitle
\begin{abstract}
    Open Domain Question Answering (QA) is evolving from complex pipelined systems to end-to-end deep neural networks. Specialized neural models have been developed for extracting answers from either text alone or Knowledge Bases (KBs) alone. In this paper we look at a more practical setting, namely QA over the combination of a KB and entity-linked text, which is appropriate when an \textit{incomplete} KB is available with a large text corpus. Building on recent advances in graph representation learning we propose a novel model, GRAFT-Net, for extracting answers from a question-specific subgraph containing text and KB entities and relations. We construct a suite of benchmark tasks for this problem, varying the difficulty of questions, the amount of training data, and KB completeness. We show that GRAFT-Net is competitive with the state-of-the-art when tested using either KBs or text alone, and vastly outperforms existing methods in the combined setting.
\end{abstract}
\section{Introduction}

Open domain Question Answering (QA) is the task of finding answers to questions posed in natural language.
Historically, this required a specialized pipeline consisting of multiple machine-learned and hand-crafted modules \citep{ferrucci2010building}.
Recently, the paradigm has shifted towards training end-to-end deep neural network models for the task \citep{chen2017reading,liang2016neural,raison2018weaver,talmor18compwebq,iyyer2017search}. Most existing models, however, answer questions using a \textit{single} information source, usually either text from an encyclopedia, or a single knowledge base (KB).

\begin{figure*}[t]
    \centering
    \includegraphics[width=0.48\linewidth,trim={2.5cm 3mm 2cm 0}, clip]{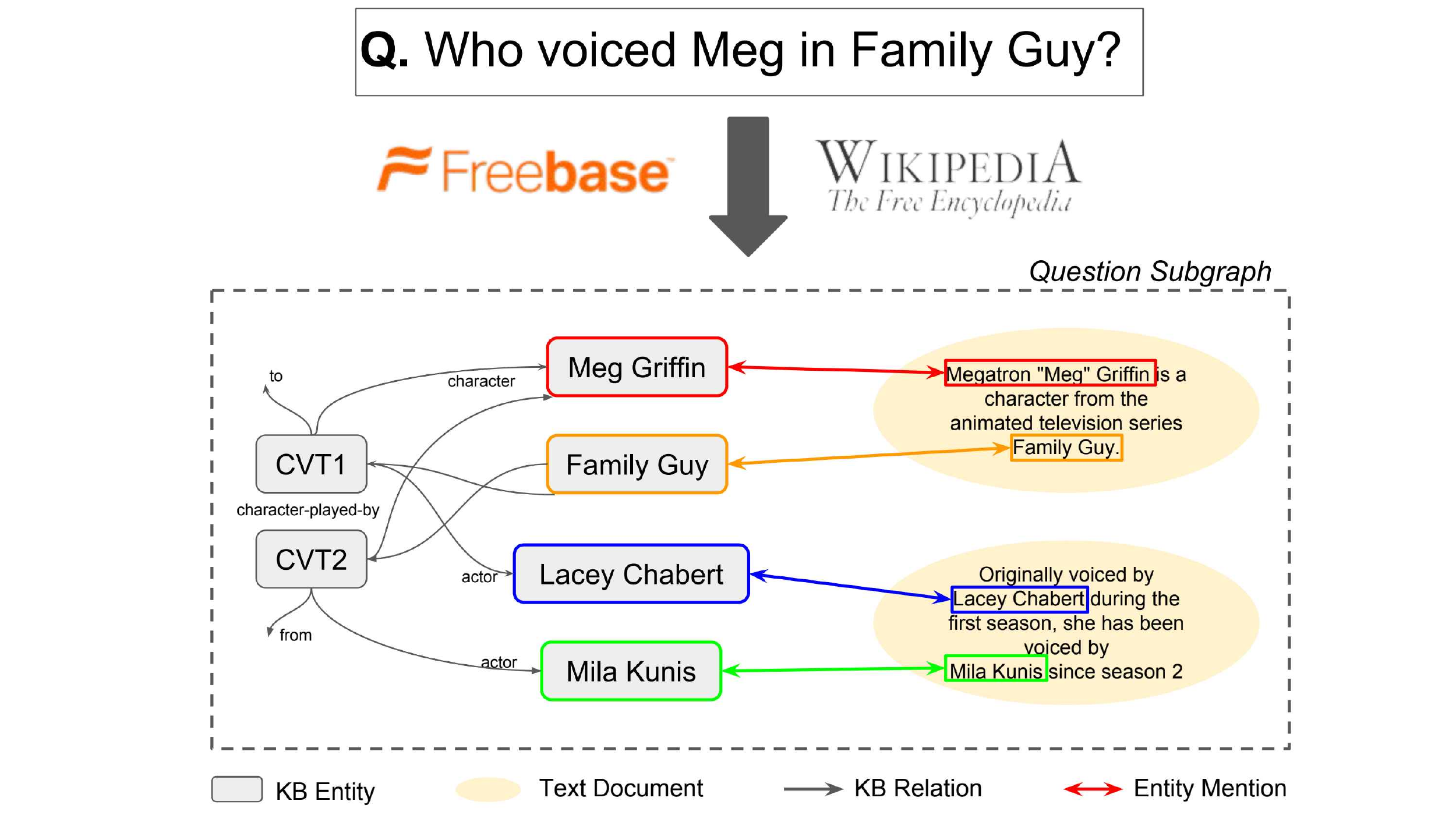}
    \raisebox{0.0\height}{\includegraphics[width=0.49\linewidth,trim={0cm 0mm 3.5cm 0cm}, clip]{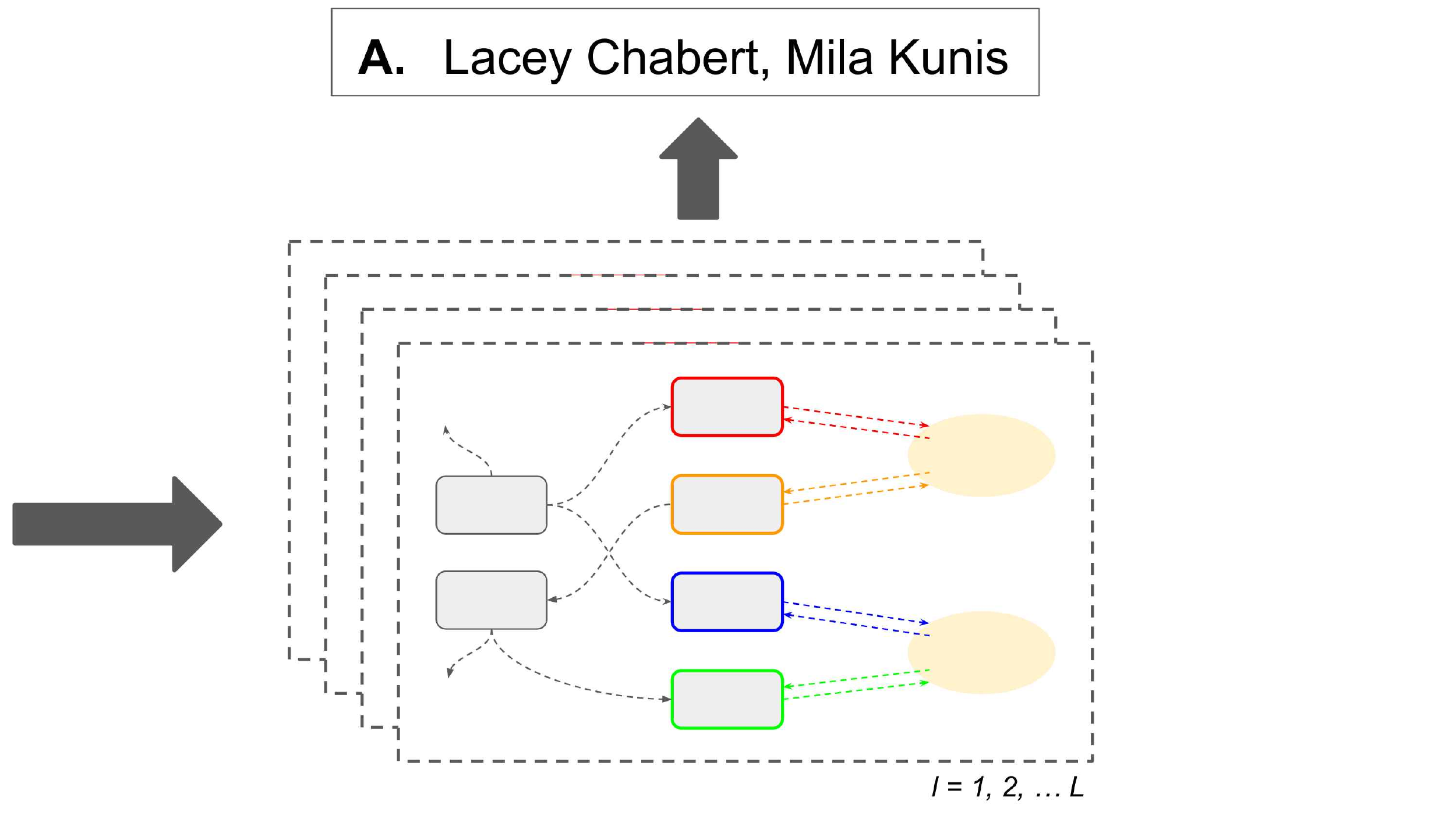}}
    \caption{\textbf{Left:} To answer a question posed in natural language, GRAFT-Net considers a heterogeneous graph constructed from text and KB facts, and thus can leverage the rich relational structure between the two information sources. \textbf{Right:} Embeddings are propagated in the graph for a fixed number of layers ($L$) and the final node representations are used to classify answers.}
    \label{fig:question-subgraph}
\end{figure*}

Intuitively, the suitability of an information source for QA depends on both its \emph{coverage} and the \emph{difficulty} of extracting answers from it. A large text corpus has high coverage, but the information is expressed using many different text patterns. 
As a result, models which operate on these patterns (e.g. BiDAF \citep{seo2016bidirectional}) do not generalize beyond their training domains \citep{wiese-weissenborn-neves:2017:CoNLL,dhingra2018simple}
or to novel types of reasoning \citep{welbl2017constructing,talmor18compwebq}.
KBs, on the other hand, suffer from low coverage due to their inevitable incompleteness and restricted schema \citep{min2013distant}, but are easier to extract answers from, since they are constructed precisely for the purpose of being queried.

In practice, 
some questions are best answered using text, while others are best answered using KBs. A natural question, then, is how to effectively combine both types of information.
Surprisingly little prior work has looked at this problem. In this paper we focus on a scenario in which a large-scale KB \citep{bollacker2008freebase,auer2007dbpedia} and a text corpus are available, but neither is sufficient alone for answering all questions.

A na\"ive option, in such a setting, is to take state-of-the-art QA systems developed for each source, and aggregate their predictions using some heuristic \citep{ferrucci2010building,baudivs2015yodaqa}. We call this approach  \emph{late fusion}, and 
show that it can be sub-optimal, as models have limited ability to aggregate evidence across the different sources (\S~\ref{sec:eval-sota-comp}). 
Instead, we focus on an \emph{early fusion} strategy, where a single model is trained to extract answers from a \emph{question subgraph} (see Fig~\ref{fig:question-subgraph}, left) containing relevant KB facts as well as text sentences. Early fusion allows
more flexibility in combining information from multiple sources. 

To enable early fusion, in this paper we propose a novel graph convolution based neural network, called GRAFT-Net (Graphs of Relations Among Facts and Text Networks), specifically designed to operate over heterogeneous graphs of KB facts and text sentences. We build upon recent work on graph representation learning \citep{kipf2016semi,schlichtkrull2017modeling}, but propose two key modifications to adopt them for the task of QA. First, we propose \emph{heterogeneous update rules} that handle KB nodes differently from the text nodes: for instance, LSTM-based updates are used to propagate information into and out of text nodes 
 (\S~\ref{sec:heterogeneous-updates}). Second, we introduce a \emph{directed propagation method}, inspired by personalized Pagerank in IR \citep{haveliwala2002topic}, which constrains the propagation of embeddings in the graph to follow paths starting from seed nodes linked to the question (\S~\ref{sec:query-conditioning}). 
Empirically, we show that both these extensions are crucial for the task of QA. An overview of the model is shown in Figure~\ref{fig:question-subgraph}.

We evaluate these methods on a new suite of benchmark tasks for testing QA models  
when both KB and text are present.
Using WikiMovies \citep{miller2016key} and WebQuestionsSP \citep{yih2016value}, we construct datasets with a varying amount of training supervision and KB completeness, and with a varying degree of question complexity. We report baselines for future comparison, including Key Value Memory Networks \citep{miller2016key,das2017question}, and show that our proposed GRAFT-Nets have superior performance across a wide range of conditions (\S~\ref{sec:experiments}). We also show that GRAFT-Nets are competitive with the state-of-the-art methods developed specifically for text-only QA, and state-of-the art methods developed for KB-only QA  (\S~\ref{sec:eval-sota-comp})\footnote{Source code and data are available at \url{https://github.com/OceanskySun/GraftNet}}.
\section{Task Setup}
\label{sec:task}

\subsection{Description}

A knowledge base is denoted as $\mathcal{K}=(\mathcal{V}, \mathcal{E}, \mathcal{R})$, where $\mathcal{V}$ is the set of entities in the KB, and the edges $\mathcal{E}$ are triplets $(s, r, o)$ which denote that relation $r\in \mathcal{R}$ holds between the subject $s\in \mathcal{V}$ and object $o\in \mathcal{V}$. A text corpus $\mathcal{D}$ is a set of documents $\{d_1, \ldots, d_{|\mathcal{D}|}\}$ where each document is a sequence of words $d_i = (w_1, \ldots, w_{|d_i|})$. We further assume that an (imperfect) entity linking system has been run on the collection of documents whose output is a set $\mathcal{L}$ of links $(v, d_p)$
connecting an entity $v\in\mathcal{V}$ with a word at position $p$ in document $d$,
and we denote with $\mathcal{L}_d$ the set of all entity links in document $d$. For entity mentions spanning multiple words in $d$, we include links to all the words in the mention in $\mathcal{L}$.

The task is, given a natural language question $q = (w_1,\ldots,w_{|q|})$, extract its answers $\{a\}_q$ from $\mathcal{G}=(\mathcal{K}, \mathcal{D}, \mathcal{L})$. There may be multiple correct answers for a question. In this paper, we assume that the answers are entities from either the documents or the KB. We are interested in a wide range of settings, where the KB $\mathcal{K}$ varies from highly incomplete to complete for answering the questions, and we will introduce datasets for testing our models under these settings.

To solve this task we proceed in two steps. First, we extract a subgraph $\mathcal{G}_q \subset \mathcal{G}$ which contains the answer to the question with high probability. The goal for this step is to ensure high recall for answers while producing a graph small enough to fit into GPU memory for gradient-based learning. Next, we use our proposed model GRAFT-Net to learn node representations in $\mathcal{G}_q$, conditioned on $q$, which are used to classify each node as being an answer or not. Training data for the second step is generated using distant supervision. The entire process mimics the search-and-read paradigm for text-based QA \citep{dhingra2017quasar}.

\subsection{Question Subgraph Retrieval}
We retrieve the subgraph $\mathcal{G}_q$ using two parallel pipelines -- one over the KB $\mathcal{K}$ which returns a set of entities, and the other over the corpus $\mathcal{D}$ which returns a set of documents. The retrieved entities and documents are then combined with entity links to produce a fully-connected graph.

\paragraph{KB Retrieval.} To retrieve relevant entities from the KB we first perform entity linking on the question $q$, producing a set of \textit{seed entities}, denoted $S_q$.
Next we run the Personalized PageRank (PPR) method \citep{haveliwala2002topic} around these seeds to identify other entities which might be an answer to the question. The edge-weights around $S_q$ are distributed equally among all edges of the same type, and they are weighted such that edges relevant to the question receive a higher weight than those which are not. Specifically, we average word vectors to compute a relation vector $v(r)$ from the surface form of the relation, and a question vector $v(q)$ from the words in the question, and use cosine similarity between these as the edge weights. After running PPR we retain the top $E$ entities $v_1,\ldots,v_E$ by PPR score, along with any edges between them, and add them to $\mathcal{G}_q$.

\paragraph{Text Retrieval.} We use Wikipedia as the corpus and retrieve text at the sentence level, i.e. documents in $\mathcal{D}$ are defined along sentences boundaries\footnote{The term \textit{document} will always refer to a sentence in the rest of this paper.}. We perform text retrieval in two steps: first we retrieve the top 5 most relevant Wikipedia articles, using the weighted bag-of-words model from DrQA \citep{chen2017reading}; then we populate a Lucene\footnote{\url{https://lucene.apache.org/}} index with sentences from these articles, and retrieve the top ranking ones $d_1,\ldots,d_D$, based on the words in the question. For the sentence-retrieval step, we found it beneficial to include the title of the article 
as an additional field in the Lucene index. As most sentences in an article talk about the title entity, this helps in retrieving relevant sentences that do not explicitly mention the entity in the question. We add the retrieved documents, along with any entities linked to them, to the subgraph $\mathcal{G}_q$.

The final question subgraph is $\mathcal{G}_q=(\mathcal{V}_q,\mathcal{E}_q,\mathcal{R}^+)$, where the vertices $\mathcal{V}_q$ consist of all the retrieved entities and documents, i.e. $\mathcal{V}_q=\{v_1,\ldots,v_E\}$ $\cup \{d_1,\ldots,d_D\}$. The edges are all relations from $\mathcal{K}$ among these entities, plus the entity-links between documents and entities, i.e.
\begin{align*}
    \mathcal{E}_q = &\{(s,o,r) \in \mathcal{E}: s,o\in \mathcal{V}_q, r\in \mathcal{R}\}\\
    &\cup\{(v,d_p,r_L):(v,d_p)\in \mathcal{L}_d, d\in\mathcal{V}_q\},
\end{align*}
where $r_L$ denotes a special ``linking'' relation. $\mathcal{R}^+ = \mathcal{R} \cup \{r_L\}$ is the set of all edge types in the subgraph.
\section{GRAFT-Nets}
\label{sec:model}

The question $q$ and its answers $\{a\}_q$ induce a labeling of the nodes in $\mathcal{V}_q$: we let $y_v=1$ if $v\in \{a\}_q$ and $y_v=0$ otherwise for all $v\in\mathcal{V}_q$ . The task of QA then reduces to performing binary classification over the nodes of the graph $\mathcal{G}_q$. Several graph-propagation based models have been proposed in the literature which learn node representations and then perform classification of the nodes \citep{kipf2016semi,schlichtkrull2017modeling}. Such models follow the standard gather-apply-scatter paradigm to learn the node representation with homogeneous updates, i.e. treating all neighbors equally.

The basic recipe for these models is as follows:
\begin{enumerate}
    \item Initialize node representations $h_v^{(0)}$.
    \item For $l=1,\ldots,L$ update node representations
    \begin{equation*}
        h_v^{(l)}=\phi\left(h_v^{(l-1)}, \sum_{v' \in N_r(v)}h_{v'}^{(l-1)}\right),
    \end{equation*}
    where $N_r(v)$ denotes the neighbours of $v$ along incoming edges of type $r$, and $\phi$ is a neural network layer.
\end{enumerate}
Here $L$ is the number of \textit{layers} in the model and corresponds to the maximum length of the paths along which information should be propagated in the graph. Once the propagation is complete the final layer representations $h_v^{(L)}$ are used to perform the desired task, for example link prediction in knowledge bases \citep{schlichtkrull2017modeling}.

However, there are two differences in our setting from previously studied graph-based classification tasks. The first difference is that, in our case, the graph $\mathcal{G}_q$ consists of \emph{heterogeneous} nodes. Some nodes in the graph correspond to KB entities which represent symbolic objects, whereas other nodes represent textual documents which are variable length sequences of words. 
The second difference is that we want to condition the representation of nodes in the graph on the natural language question $q$. In \S \ref{sec:heterogeneous-updates} we introduce heterogeneous updates to address the first difference, and in \S \ref{sec:query-conditioning} we introduce mechanisms for conditioning on the question (and its entities) for the second.

\subsection{Node Initialization}
Nodes corresponding to entities are initialized using fixed-size vectors $h_v^{(0)}=x_v \in \mathbb{R}^n$, where $x_v$ can be pre-trained KB embeddings or random, and $n$ is the embedding size. Document nodes in the graph describe a variable length sequence of text. Since multiple entities might link to different positions in the document, we maintain a variable length representation of the document in each layer. This is denoted by $H_{d}^{(l)}\in \mathbb{R}^{|d|\times n}$. Given the words in the document $(w_1,\ldots,w_{|d|})$, we initialize its hidden representation as:
\begin{equation*}
    H_d^{(0)} = \text{LSTM}(w_1, w_2, \dots ),
\end{equation*}
where LSTM refers to a long short-term memory unit.
We denote the $p$-th row of $H_d^{(l)}$, corresponding to the embedding of $p$-th word in the document $d$ at layer $l$, as $H_{d,p}^{(l)}$. 

\begin{figure*}[!t]
    \centering
    \includegraphics[width=0.8\textwidth]{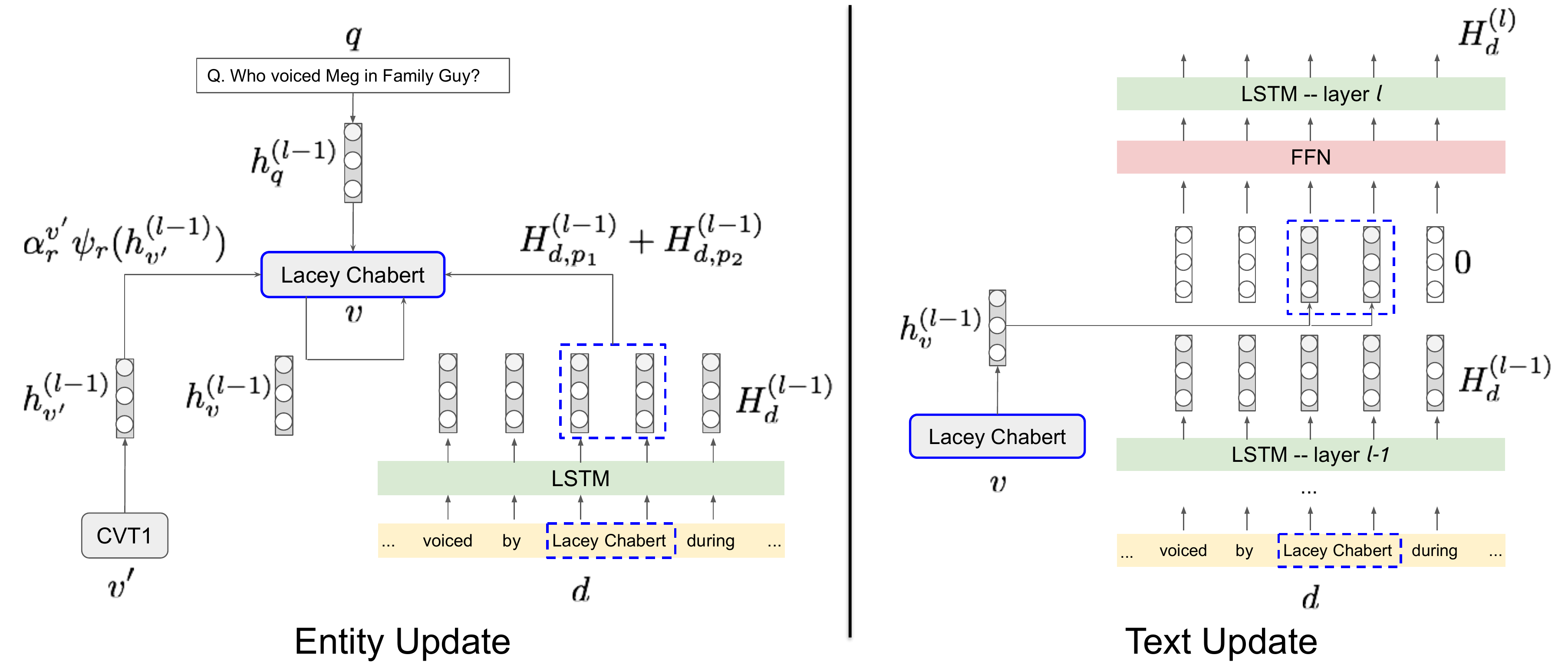}
    \caption{Illustration of the heterogeneous update rules for entities (\textbf{left}) and text documents (\textbf{right})}
    \label{fig:entity_update}
\end{figure*}

\subsection{Heterogeneous Updates}
\label{sec:heterogeneous-updates}
Figure~\ref{fig:entity_update} shows the update rules for entities and documents, which we describe in detail here.

\paragraph{Entities.}
Let $M(v)=\{(d,p)\}$ be the set of positions $p$ in documents $d$ which correspond to a mention of entity $v$.
The update for entity nodes involves a single-layer feed-forward network (FFN) over the concatenation of four states:
\begin{equation}
    h_v^{(l)} = \text{FFN}\left(
    \begin{bmatrix}
        h_v^{(l-1)} \\
        h_q^{(l-1)} \\
        \sum_r \sum_{v'\in N_r(v)} \alpha_{r}^{v'} \psi_r(h_{v'}^{(l-1)}) \\
        \sum_{(d,p) \in M(v)} H_{d,p}^{(l-1)}
    \end{bmatrix}
    \right).
    \label{eq:entity-upd}
\end{equation}
The first two terms correspond to the entity representation and question representation (details below), respectively, from the previous layer.

The third term aggregates the states from the entity neighbours of the current node, $N_r(v)$, after scaling with an attention weight $\alpha_{r}^{v'}$ (described in the next section), and applying relation specific transformations $\psi_r$. 
Previous work on Relational-Graph Convolution Networks \citep{schlichtkrull2017modeling} used a linear projection for $\psi_r$. For a batched implementation, this results in matrices of size $O(B|\mathcal{R}_q||\mathcal{E}_q|n)$, where $B$ is the batch size,
which can be prohibitively large for large subgraphs\footnote{This is because we have to use adjacency matrices of size $|\mathcal{R}_q|\times|\mathcal{E}_q|\times|\mathcal{E}_q|$ to aggregate embeddings from neighbours of all nodes simultaneously.}. Hence in this work we use \textit{relation vectors} $x_r$ for $r\in\mathcal{R}_q$ instead of matrices, and compute the update along an edge as:
\begin{equation}
    \psi_r(h_{v'}^{(l-1)}) = pr_{v'}^{(l-1)}\text{FFN}\left( x_r, h_{v'}^{(l-1)}\right).
    \label{eq:edge-upd}
\end{equation}
Here $pr_{v'}^{(l-1)}$ is a PageRank score used to control the propagation of embeddings along paths starting from the seed nodes, which we describe in detail in the next section.
The memory complexity of the above is $O(B(|\mathcal{F}_q|+|\mathcal{E}_q|)n)$, where $|\mathcal{F}_q|$ is the number of facts in the subgraph $\mathcal{G}_q$.

The last term aggregates the states of all tokens that correspond to mentions of the entity $v$ among the documents in the subgraph. Note that the update depends on the positions of entities in their containing document.



\paragraph{Documents.} 
Let $L(d, p)$ be the set of all entities linked to the word at position $p$ in document $d$. 
The document update proceeds in two steps. First we aggregate over the entity states coming in at each position separately:
\begin{subequations}
\begin{equation}
    \Tilde{H}_{d,p}^{(l)} = \text{FFN}\left(H_{d,p}^{(l-1)}, \sum_{v\in L(d,p)}h_v^{(l-1)}\right).
    \label{eq:doc-upd-a}
\end{equation}
Here $h_v^{(l-1)}$ are normalized by the number of outgoing edges at $v$. Next we aggregate states within the document using an LSTM:
\begin{equation}
    H_d^{(l)} = \text{LSTM}(\Tilde{H}_d^{(l)}).
    \label{eq:doc-upd-b}
\end{equation}
\end{subequations}

\subsection{Conditioning on the Question}
\label{sec:query-conditioning}
For the parts described thus far, the graph learner is largely agnostic of the question. We introduce dependence on question in two ways: by attention over relations, and by personalized propagation. 

To represent $q$, let $w_1^q,\ldots,w_{|q|}^q$ be the words in the question. The initial representation is computed as:
\begin{equation}
    h_q^{(0)} = \text{LSTM}(w_1^q,\ldots,w_{|q|}^q)_{|q|} \in \mathbb{R}^n,
\end{equation}
where we extract the final state from the output of the LSTM. In subsequent layers the question representation is updated as $h_q^{(l)} = \text{FFN}\left(\sum_{v\in S_q}h_v^{(l)} \right)$, where $S_q$ denotes the seed entities mentioned in the question.

\paragraph{Attention over Relations.}
The attention weight in the third term of Eq. (\ref{eq:entity-upd}) is computed using the question and relation embeddings:
\begin{equation*}
    \alpha_{r}^{v'} = \text{softmax}(x_r^T h_q^{(l-1)}),
\end{equation*}
where the softmax normalization is over all outgoing edges from $v'$, and $x_r$ is the relation vector for relation $r$. This ensures that embeddings are propagated more along edges relevant to the question.

\begin{figure}[t]
    \centering
    \includegraphics[width=0.7\linewidth,trim={3cm 2.5cm 3cm 2.5cm}, clip]{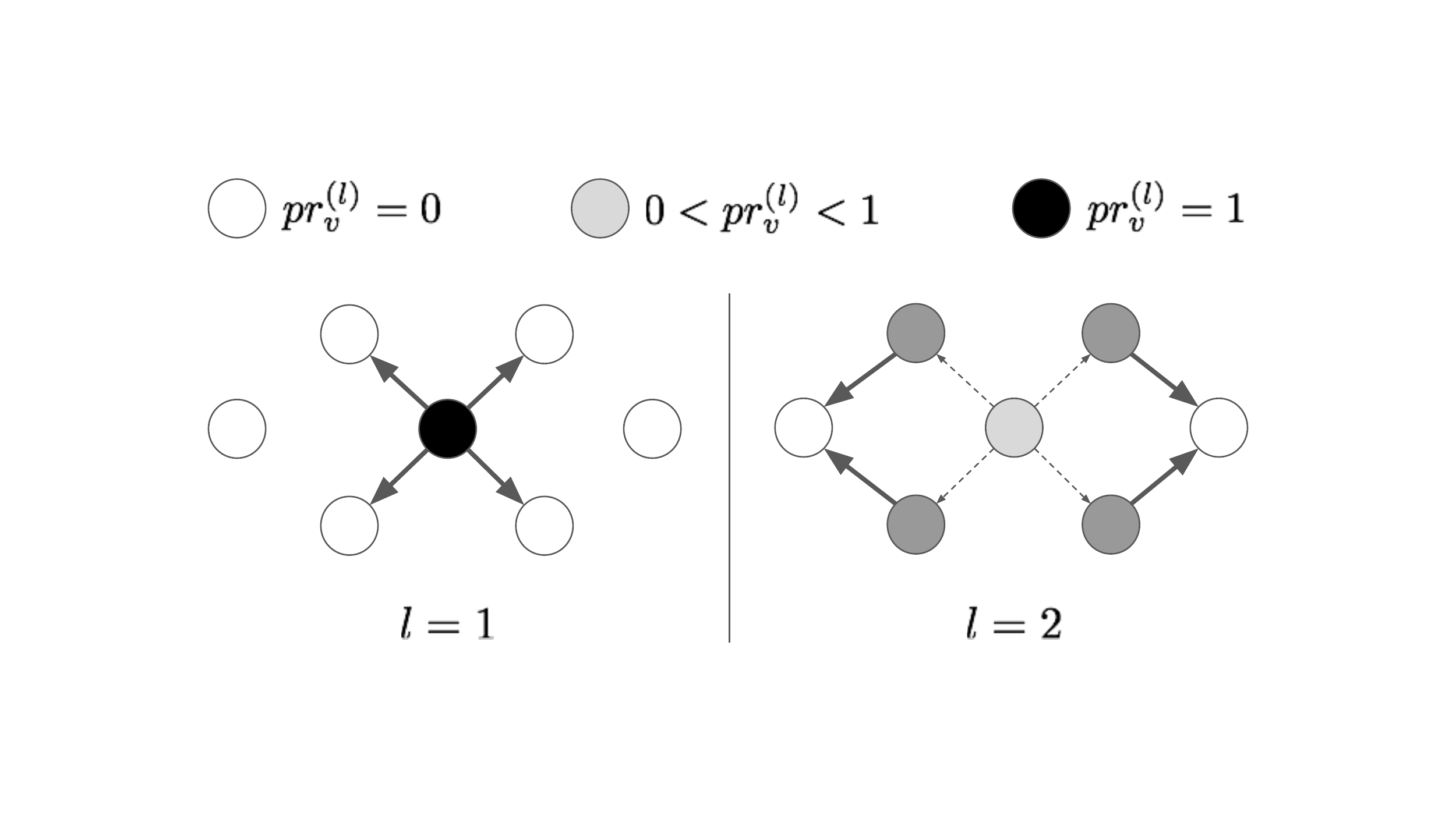}
    \caption{Directed propagation of embeddings in GRAFT-Net. A scalar \textit{PageRank} score $pr_v^{(l)}$ is maintained for each node $v$ across layers, which spreads out from the seed node. Embeddings are only propagated from nodes with $pr_v^{(l)}>0$.}
    \label{fig:pagerank_prop}
\end{figure}

\paragraph{Directed Propagation.}
Many questions require multi-hop reasoning, which follows a path from a seed node mentioned in the question to the target answer node. To encourage such a behaviour when propagating embeddings, we develop a technique inspired from personalized PageRank in IR \citep{haveliwala2002topic}. The propagation starts at the seed entities $S_q$ mentioned in the question. In addition to the vector embeddings $h_v^{(l)}$ at the nodes, we also maintain scalar ``PageRank'' scores $pr_v^{(l)}$ which measure the total weight of paths from a seed entity to the current node, as follows:

\begin{align*}
    pr_v^{(0)} &= \begin{cases}
        \frac{1}{|S_q|} \quad &\text{if}\quad v \in S_q\\
        0 \quad &\text{o.w.}
    \end{cases},\\
    pr_v^{(l)} &= (1-\lambda)pr_v^{(l-1)} + \lambda\sum_r\sum_{v'\in N_r(v)} \alpha_{r}^{v'} pr_{v'}^{(l-1)}.
\end{align*}
Notice that we reuse the attention weights $\alpha_{r}^{v'}$ when propagating PageRank, to ensure that nodes along paths relevant to the question receive a high weight. The PageRank score is used as a scaling factor when propagating embeddings along the edges in Eq. (\ref{eq:edge-upd}). For $l=1$, the PageRank score will be $0$ for all entities except the seed entities, and hence propagation will only happen outward from these nodes. For $l=2$, it will be non-zero for the seed entities and their 1-hop neighbors, and propagation will only happen along these edges. Figure~\ref{fig:pagerank_prop} illustrates this process.

\subsection{Answer Selection}
The final representations $h^{(L)}_v \in \mathbb{R}^n$, are used for binary classification to select the answers:
\begin{equation}
    \Pr\left(v \in \{a\}_q|\mathcal{G}_q, q\right) = \sigma(w^T h^{(L)}_v + b),
\end{equation}
where $\sigma$ is the sigmoid function. Training uses binary cross-entropy loss over these probabilities.

\subsection{Regularization via Fact Dropout}
To encourage the model to learn a robust classifier, which exploits all available sources of information, we randomly drop edges from the graph during training with probability $p_0$. We call this \textit{fact-dropout}. It is usually easier to extract answers from the KB than from the documents, so the model tends to rely on the former, especially when the KB is complete. This method is similar to DropConnect \citep{wan2013regularization}.

\section{Related Work}
The work of \citet{das2017question} attempts an early fusion strategy for QA over KB facts and text. Their approach is based on Key-Value Memory Networks (KV-MemNNs) \citep{miller2016key} coupled with a universal schema \citep{riedel2013relation} to populate a memory module with representations of KB triples and text snippets independently. The key limitation for this model is that it ignores the rich relational structure between the facts and text snippets. Our graph-based method, on the other hand, explicitly uses this structure for the propagation of embeddings. We compare the two approaches in our experiments (\S \ref{sec:experiments}), and show that GRAFT-Nets outperform KV-MemNNs over all tasks. 

Non-deep learning approaches have been also attempted for QA over both text assertions and KB facts. \citet{gardner2017open} use traditional feature extraction methods of open-vocabulary semantic parsing for the task. \citet{RYU2014683} use a pipelined system aggregating evidence from both unstructured and semi-structured sources for open-domain QA.

Another line of work has looked at learning combined representations of KBs and text for relation extraction and Knowledge Base Completion (KBC) \citep{lao2012reading,riedel2013relation,toutanova2015representing,verga2015multilingual,das2016chains,han2016joint}. 
The key difference in QA compared to KBC is that in QA the inference process on the knowledge source has to be conditioned on the question, so different questions induce different representations of the KB and warrant a different inference process.
Furthermore, KBC operates under the fixed schema defined by the KB before-hand, whereas natural language questions might not adhere to this schema.

The GRAFT-Net model itself is motivated from the large body of work on graph representation learning \citep{scarselli2009graph,li2015gated,kipf2016semi,atwood2016diffusion, schlichtkrull2017modeling}. 
Like most other graph-based models, GRAFT-Nets can also be viewed as an instantiation of the Message Passing Neural Network (MPNN) framework of \citet{gilmer2017neural}. GRAFT-Nets are also \textit{inductive} representation learners like GraphSAGE \citep{DBLP:journals/corr/HamiltonYL17}, but operate on a heterogeneous mixture of nodes and use retrieval for getting a subgraph instead of random sampling. The recently proposed Walk-Steered Convolution model uses random walks for learning graph representations \citep{jiang2018walk}. Our personalization technique also borrows from such random walk literature, but uses it to localize propagation of embeddings.


Tremendous progress on QA over KB has been made with deep learning based approaches like memory networks \citep{bordes2015large,jain2016question} and reinforcement learning \citep{liang2016neural, das2017go}. But extending them with text, which is our main focus, is non-trivial. In another direction, there is also work on producing parsimonious graphical representations of textual data \citep{krause2016sar,lu2017object}; however in this paper we use a simple sequential representation augmented with entity links to the KB which works well.

For QA over text only, a major focus has been on the task of reading comprehension \citep{seo2016bidirectional,gong2017ruminating,hu2017mnemonic,shen2017reasonet,yu2018qanet} since the introduction of SQuAD \citep{rajpurkar2016squad}. 
These systems assume that the answer-containing passage is known apriori, but there has been progress when this assumption is relaxed 
\citep{chen2017reading,raison2018weaver,dhingra2017quasar,wang2017r,wang2017evidence,watanabe2017question}. We work in the latter setting, where relevant information must be retrieved from large information sources, but we also incorporate KBs into this process.

\section{Experiments \& Results}
\label{sec:experiments}



\subsection{Datasets}
\textbf{WikiMovies-10K} consists of $10K$ randomly sampled training questions from the WikiMovies dataset \citep{miller2016key}, along with the original test and validation sets. We sample the training questions to create a more difficult setting, since the original dataset has $100K$ questions over only $8$ different relation types, which is unrealistic in our opinion. In \S~\ref{sec:eval-sota-comp} we also compare to the existing state-of-the-art using the full training set.

We use the KB and text corpus constructed from Wikipedia released by \citet{miller2016key}. For entity linking we use simple surface level matches, and retrieve the top $50$ entities around the seeds to create the question subgraph. We further add the top $50$ sentences (along with their article titles) to the subgraph using Lucene search over the text corpus. The overall answer recall in our constructed subgraphs is $99.6\%$.

\noindent \textbf{WebQuestionsSP} \citep{yih2016value} consists of $4737$ natural language questions posed over Freebase entities, split up into $3098$ training and $1639$ test questions. We reserve $250$ training questions for model development and early stopping. We use the entity linking outputs from S-MART\footnote{\url{https://github.com/scottyih/STAGG}} and retrieve $500$ entities from the neighbourhood around the question seeds in Freebase to populate the question subgraphs\footnote{A total of $13$ questions had no detected entities. These were ignored during training and considered as incorrect during evaluation.}. We further retrieve the top $50$ sentences from Wikipedia with the two-stage process described in \S \ref{sec:task}. The overall recall of answers among the subgraphs is $94.0\%$.

\begin{table*}[!htbp]
\small
\centering
\begin{tabular}{@{}cccccc@{}}
\toprule
\textbf{Dataset} & \textbf{\# train / dev / test} & \textbf{\# entity nodes} & \textbf{\# edge types} & \textbf{\# document nodes} & \textbf{\# question vocab} \\ \midrule
WikiMovies-10K   & 10K / 10K / 10K                & 43,233               & 9                     & 79,728             & 1759                    \\
WebQuestionsSP   & 2848 / 250 / 1639              & 528,617              & 513                   & 235,567            & 3781                    \\ \bottomrule
\end{tabular}
\caption{Statistics of all the retrieved subgraphs $\cup_q \mathcal{G}_q$ for WikiMovies-10K and WebQuestionsSP. \label{tab:datasets}}
\end{table*}

Table~\ref{tab:datasets} shows the combined statistics of all the retreived subgraphs for the questions in each dataset.  These two datasets present varying levels of difficulty. While all questions in WikiMovies correspond to a single KB relation, for WebQuestionsSP the model needs to aggregate over two KB facts for $\sim$$30\%$ of the questions, and also requires reasoning over constraints for $\sim$$7\%$ of the questions \citep{liang2016neural}.
For maximum portability, QA systems need to be robust across several degrees of KB availability since  different domains might contain different amounts of structured data; and KB completeness may also vary over time.
Hence, we construct an additional $3$ datasets each from the above two, with the number of KB facts downsampled to $10\%$, $30\%$ and $50\%$ of the original to simulate settings where the KB is incomplete. We repeat the retrieval process for each sampled KB.

\begin{table*}
\begin{minipage}[bt]{0.63\linewidth}
\resizebox{\linewidth}{!}{
\centering
\small
\begin{tabular}[t]{@{}lccccc@{}}
\toprule
\multicolumn{1}{c}{\multirow{2}{*}{Model}} & \multirow{2}{*}{Text Only} & \multicolumn{4}{c}{KB + Text}                                                             \\ \cmidrule(l){3-6} 
\multicolumn{1}{c}{}                       &                            & 10 \%                & 30\%                 & 50\%                 & 100\%                \\ \midrule
\multicolumn{6}{l}{WikiMovies-10K}                                                                                                                                \\ \midrule
KV-KB                                      & --                         & 15.8 / ~9.8             & 44.7 / 30.4            & 63.8 / 46.4            & 94.3 / 76.1            \\
KV-EF                                      & 50.4 / 40.9                  & 53.6 / 44.0            & 60.6 / 48.1            & 75.3 / 59.1            & 93.8 / 81.4            \\
GN-KB                                      & --                         & 19.7 / 17.3            & 48.4 / 37.1            & 67.7 / 58.1            & \textbf{97.0 / 97.6}            \\
GN-LF                                      & \multirow{3}{*}{$\left. \vphantom{\begin{tabular}{c}3\\3\\3\end{tabular}}\right\{$\textbf{73.2 / 64.0}$\left. \vphantom{\begin{tabular}{c}3\\3\\3\end{tabular}}\right\}$} & 74.5 / 65.4          & 78.7 / 68.5          & 83.3 / 74.2          & 96.5 / 92.0            \\
GN-EF                                      &                            & 75.4 / 66.3          & 82.6 / 71.3          & 87.6 / 76.2          & 96.9 / 94.1          \\
GN-EF+LF                                   &                            & \textbf{79.0 / 66.7}            & \textbf{84.6 / 74.2}          & \textbf{88.4 / 78.6}          & \textbf{96.8 / 97.3}          \\ \midrule
\multicolumn{6}{l}{WebQuestionsSP}                                                                                                                                  \\ \midrule
KV-KB                                      & --                         & 12.5 / ~4.3             & 25.8 / 13.8            & 33.3 / 21.3            & 46.7 / 38.6            \\
KV-EF                                      & 23.2 / 13.0                & 24.6 / 14.4          & 27.0 / 17.7            & 32.5 / 23.6            & 40.5 / 30.9            \\
GN-KB                                      & --                         & 15.5 / ~6.5             & 34.9 / 20.4            & 47.7 / 34.3            & 66.7 / 62.4            \\
GN-LF                                      & \multirow{3}{*}{$\left. \vphantom{\begin{tabular}{c}3\\3\\3\end{tabular}}\right\{$\textbf{25.3 / 15.3}$\left. \vphantom{\begin{tabular}{c}3\\3\\3\end{tabular}}\right\}$}
& 29.8 / 17.0            & 39.1 / 25.9          & 46.2 / 35.6          & 65.4 / 56.8          \\
GN-EF                                      &                            & 31.5 / 17.7          & 40.7 / 25.2          & 49.9 / 34.7          & 67.8 / 60.4          \\
GN-EF+LF                                   &                            & \textbf{33.3 / 19.3}          & \textbf{42.5 / 26.7}          & \textbf{52.3 / 37.4}          & \textbf{68.7 / 62.3}          \\ \bottomrule
\end{tabular}
}

\end{minipage}\hfill
\begin{minipage}[bt]{0.3\linewidth}
    \centering
    \includegraphics[width=1.0\linewidth]{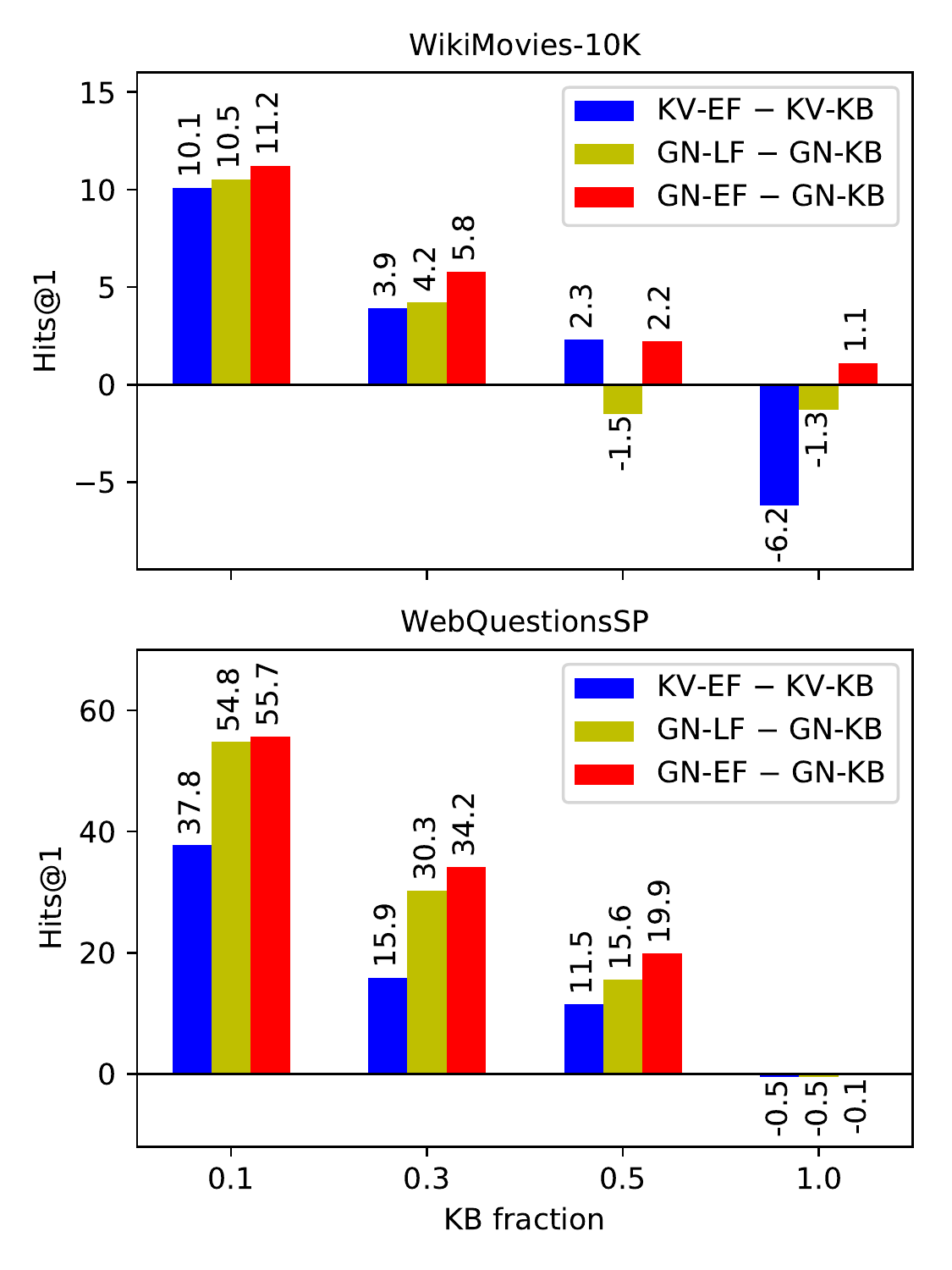}
\end{minipage}
    \caption{\textbf{Left:} Hits@1 / F1 scores of GRAFT-Nets (GN) compared to KV-MemNN (KV) in KB only (-KB), early fusion (-EF), and late fusion (-LF) settings. \textbf{Right:} Improvement of early fusion (-EF) and late fusion (-LF) over KB only (-KB) settings as KB completeness increases. \label{table:main}}
\end{table*}

\subsection{Compared Models}
\textbf{KV-KB} is the Key Value Memory Networks model from \citet{miller2016key,das2017question} but using only KB and ignoring the text. \textbf{KV-EF} (early fusion) is the same model with access to both KB and text as memories. For text we use a BiLSTM over the entire sentence as keys, and entity mentions as values. This re-implementation shows better performance on the text-only and KB-only WikiMovies tasks than the results reported previously\footnote{For all KV models we tuned the number of layers $\{1,2,3\}$, batch size $\{10, 30, 50\}$, model dimension $\{50, 80\}$. We also use fact dropout regularization in the KB+Text setting tuned between $\{0, 0.2, 0.4\}$.} (see Table~\ref{table:soa_new}). \textbf{GN-KB} is the GRAFT-Net model ignoring the text. \textbf{GN-LF} is a late fusion version of the GRAFT-Net model: we train two separate models, one using text only and the other using KB only, and then ensemble the two\footnote{For ensembles we take a weighted combination of the answer probabilities produced by the models, with the weights tuned on the dev set. For answers only in text or only in KB, we use the probability as is.}. \textbf{GN-EF} is our main GRAFT-Net model with early fusion. \textbf{GN-EF+LF} is an ensemble over the GN-EF and GN-LF models, with the same ensembling method as GN-LF. 
We report Hits@1, which is the accuracy of the top-predicted answer from the model, and the F1 score. To compute the F1 score we tune a threshold on the development set to select answers based on binary probabilities for each node in the subgraph.

\subsection{Main Results}
Table~\ref{table:main} presents a comparison of the above models across all datasets.
GRAFT-Nets (GN) shows consistent improvement over KV-MemNNs on both datasets in all settings, including KB only (-KB), text only (-EF, Text Only column), and early fusion (-EF). Interestingly, we observe a larger relative gap between the Hits and F1 scores for the KV models than we do for our GN models. We believe this is because the attention for KV is normalized over the memories, which are KB facts (or text sentences): hence the model is unable to assign high probabilities to multiple facts at the same time. On the other hand, in GN, we normalize the attention over \textit{types of relations} outgoing from a node, and hence can assign high weights to all the correct answers.

We also see a consistent improvement of early fusion over late fusion (-LF), and by ensembling them together we see the best performance across all the models. In Table~\ref{table:main} (right), we further show the improvement for KV-EF over KV-KB, and GN-LF and GN-EF over GN-KB, as the amount of KB is increased. This measures how effective these approaches are in utilizing text plus a KB. For KV-EF we see improvements when the KB is highly incomplete, but in the full KB setting, the performance of the fused approach is worse. A similar trend holds for GN-LF. On the other hand, GN-EF with text improves over the KB-only approach in all settings. As we would expect, though, the benefit of adding text decreases as the KB becomes more and more complete.

\subsection{Comparison to Specialized Methods}
\label{sec:eval-sota-comp}
In Table~\ref{table:soa_new} we compare GRAFT-Nets to state-of-the-art models that are specifically designed and tuned for QA using either only KB or only text. For this experiment we use the full WikiMovies dataset to enable direct comparison to previously reported numbers. For DrQA \citep{chen2017reading}, following the original paper, we restrict answer spans for WebQuestionsSP to match an entity in Freebase. In each case we also train GRAFT-Nets using only KB facts or only text sentences. In three out of the four cases, we find that GRAFT-Nets either match or outperform the existing state-of-the-art models. We emphasize that the latter have no mechanism for dealing with the fused setting.

\begin{table*}[!htbp]
\centering
\small
\begin{tabular}{@{}lll@{}}
\toprule
\textbf{Question}                                 & \textbf{Correct Answers}                                                                                           & \textbf{Predicted Answers}                                                                                      \\ \midrule
what language do most people speak in afghanistan & \begin{tabular}[c]{@{}l@{}}Pashto language,\\ {\color[HTML]{FFC702} Farsi (Eastern Language)}\end{tabular} & Pashto language                                                          \\ \midrule
what college did john stockton go to              & Gonzaga University                                                                                                 & \begin{tabular}[c]{@{}l@{}}Gonzaga University,\\ {\color[HTML]{CB0000} Gonzaga Preparatory School}\end{tabular} \\ \bottomrule
\end{tabular}
\caption{Examples from WebQuestionsSP dataset. \textbf{Top:} The model {\color[HTML]{FFC702} misses} a correct answer. \textbf{Bottom:} The model predicts an extra {\color[HTML]{CB0000} incorrect} answer. \label{fig:example}}
\end{table*}

The one exception is the KB-only case for WebQuestionsSP where GRAFT-Net does $6.2\%$ F1 points worse than Neural Symbolic Machines \citep{liang2016neural}. Analysis suggested three explanations: (1) In the KB-only setting, the recall of subgraph retrieval is only $90.2\%$, which limits overall performance. In an oracle setting where we ensure the answers are part of the subgraph, the F1 score increases by $4.8\%$. (2) We use the same probability threshold for all questions, even though the number of answers may vary significantly. Models which parse the query into a symbolic form do not suffer from this problem since answers are retrieved in a deterministic fashion. If we tune separate thresholds for each question the F1 score improves by $7.6\%$. (3) GRAFT-Nets perform poorly in the few cases where there is a constraint involved in picking out the answer (for example, ``who \textit{first} voiced Meg in Family Guy''). If we ignore such constraints, and consider all entities with the same sequence of relations to the seed as correct, the performance improves by $3.8\%$ F1. Heuristics such as those used by \citet{yu2017improved} can be used to improve these cases. Figure~\ref{fig:example} shows examples where GRAFT-Net fails to predict the correct answer set exactly.

\begin{table}[t]
\resizebox{\linewidth}{!}{
\begin{tabular}{@{} lccccc @{}}
\toprule
 \multirow{2}{*}{Method}                           & \multicolumn{2}{c}{WikiMovies (full)} && \multicolumn{2}{c}{WebQuestionsSP} \\
  \cline{2-3}\cline{5-6}
  & kb            & doc   && kb           & doc         \\
\midrule
MINERVA                  & \textbf{97.0 / --~~~~~}     & --            &  & --          & --          \\
R2-AsV  & --            & 85.8 / --~~~~~     &  & --             & --             \\
NSM & --            & --            &  & \textbf{~~~~~-- / 69.0}   & --          \\
DrQA*                    & --            & --            &  & --          & 21.5 / --~~~~~   \\
R-GCN$^\#$                   & \textbf{96.5 / 97.4}   & --            &  & 37.2 / 30.5 & --          \\
KV                       & 93.9 / --~~~~~     & 76.2 / --~~~~~     &  & -- / --     & -- / --     \\
KV$^\#$                      & 95.6 / 88.0   & 80.3 / 72.1   &  & 46.7 / 38.6 & 23.2 / 13.0 \\
GN                       & \textbf{96.8 / 97.2}   & \textbf{86.6 / 80.8}   &  & 67.8 / 62.8 & \textbf{25.3 / 15.3} \\ \bottomrule
\end{tabular}}
\caption{Hits@1 / F1 scores compared to SOTA models using only KB or text: MINERVA \cite{das2017go}, R2-AsV \cite{watanabe2017question}, Neural Symbolic Machines (NSM) \cite{liang2016neural}, DrQA \cite{chen2017reading}, R-GCN \cite{schlichtkrull2017modeling} and KV-MemNN \citep{miller2016key}. *DrQA is pretrained on SQuAD. $^\#$Re-implemented. \label{table:soa_new}
}
\end{table}

\subsection{Effect of Model Components}
\paragraph{Heterogeneous Updates.}
We tested a non-heterogeneous version of our model, where instead of using fine-grained entity linking information for updating the node representations ($M(v)$ and $L(d,p)$ in Eqs. \ref{eq:entity-upd}, \ref{eq:doc-upd-a}), we aggregate the document states across all its positions as $\sum_p H_{d,p}^{(l)}$ and use this combined state for all updates.
Without the heterogeneous update, all entities $v\in L(d, \cdot)$ will receive the same update from document $d$. Therefore, the model cannot disambiguate different entities mentioned in the same document. 
The result in Table \ref{table:hetero} shows that this version is consistently worse than the heterogeneous model.

\begin{table}[htp]
\centering
\resizebox{\linewidth}{!}{
\begin{tabular}{@{}lccccc@{}}
\toprule
          & 0 KB      & 0.1 KB    & 0.3 KB    & 0.5 KB    & 1.0 KB    \\ \hline
NH & 22.7 / 13.6 & 28.7 / 15.8 & 35.6 / 23.2 & 47.2 / 33.3 & 66.5 / 59.8 \\
H     & 25.3 / 15.3 & 31.5 / 17.7 & 40.7 / 25.2 & 49.9 / 34.7 & 67.8 / 60.4 \\
\bottomrule
\end{tabular}}
\caption{Non-Heterogeneous (NH) vs. Heterogeneous (H) updates on WebQuestionsSP}
\label{table:hetero}
\end{table}

\begin{figure}[t]
    \centering
    \includegraphics[width=0.48\linewidth]{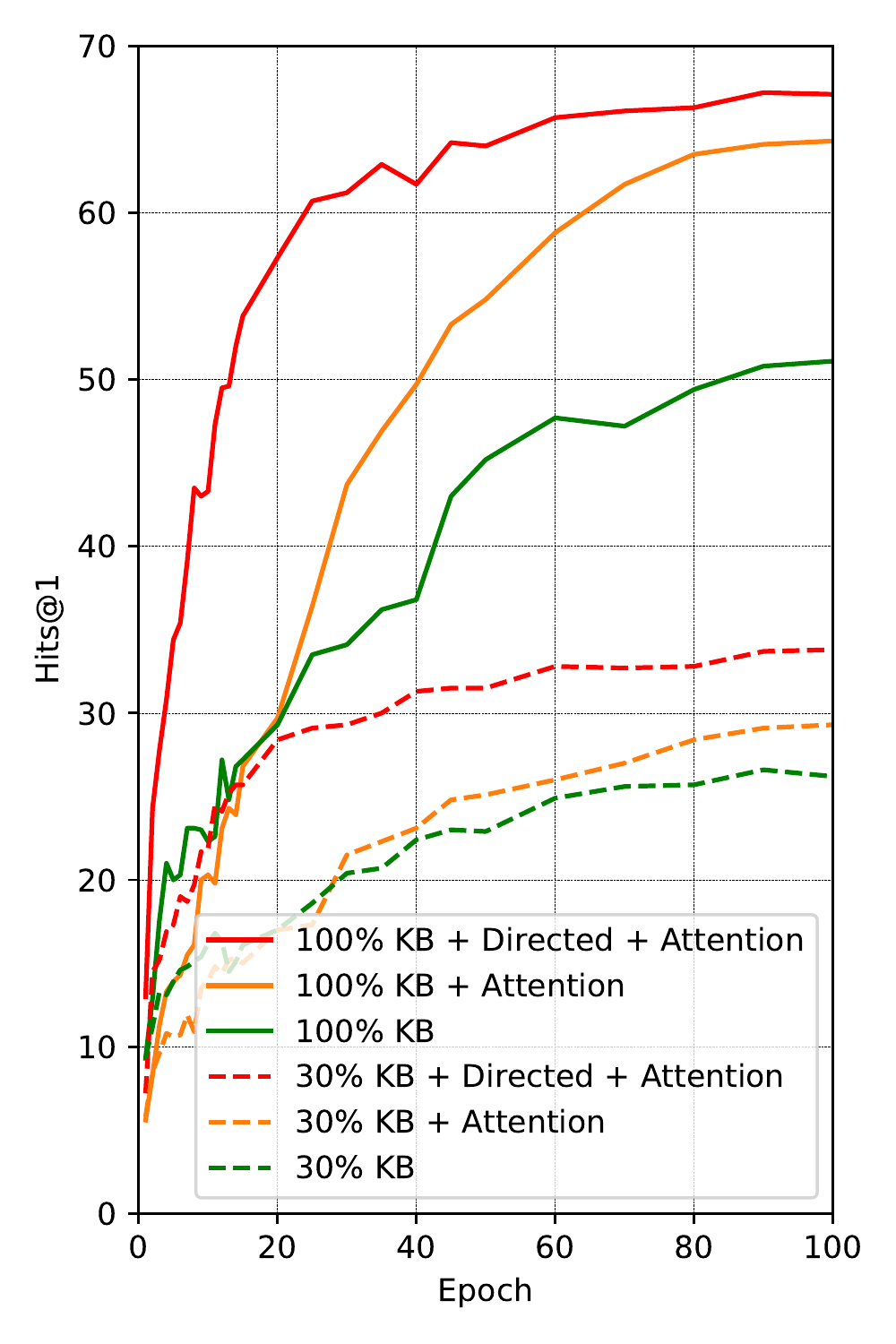}~
    \includegraphics[width=0.48\linewidth]{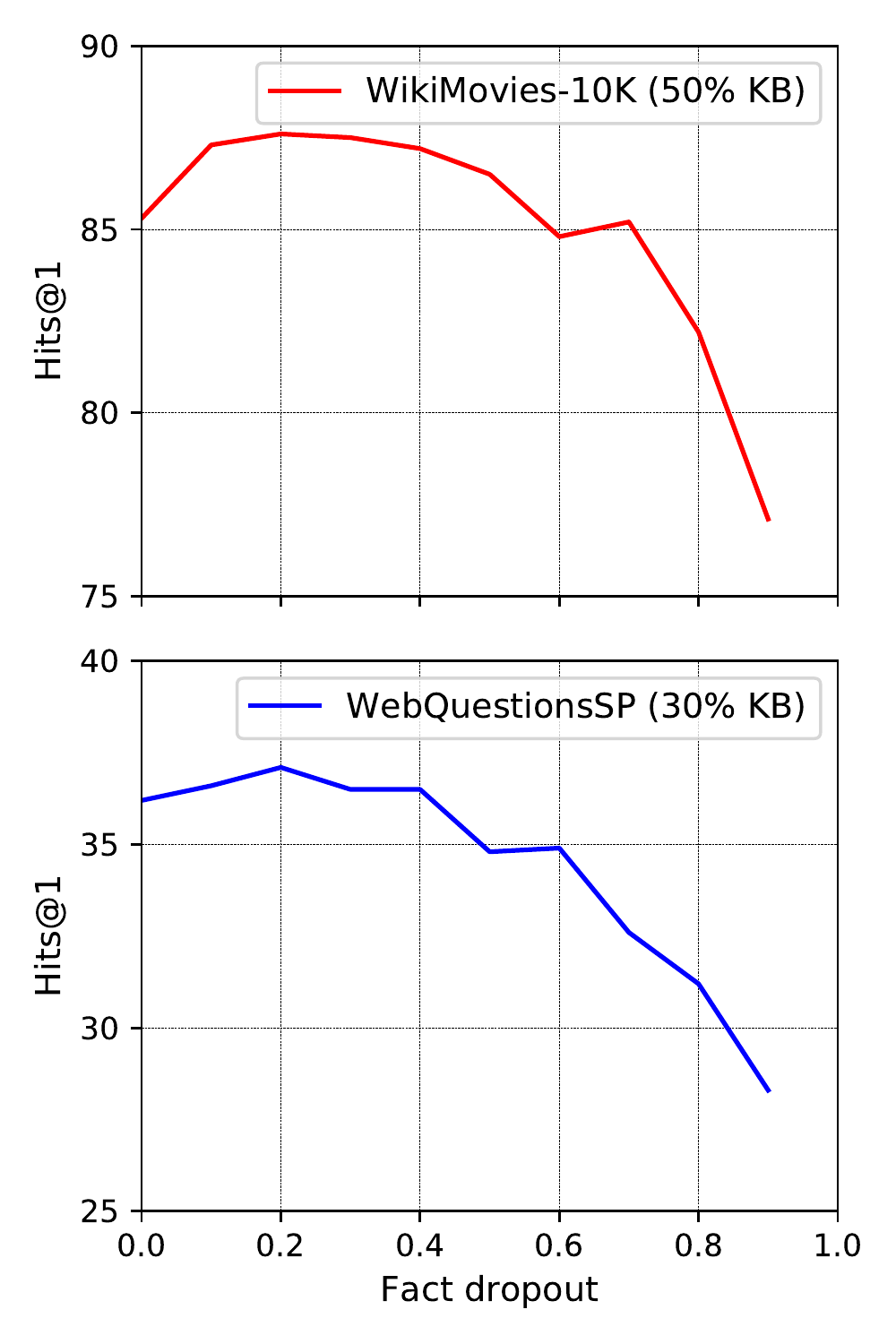}
    \caption{\textbf{Left:} Effect of directed propagation and query-based attention over relations for the WebQuestionsSP dataset with 30\% KB and 100\% KB. \textbf{Right:} Hits@1 with different rates of fact-dropout on and WikiMovies and WebQuestionsSP.}
    \label{fig:pagerank}
\end{figure}

\paragraph{Conditioning on the Question.}
We performed an ablation test on the directed propagation method and attention over relations. We observe that both components lead to better performance. Such effects are observed in both complete and incomplete KB scenarios, e.g. on WebQuestionsSP dataset, as shown in Figure \ref{fig:pagerank} (left).

\paragraph{Fact Dropout.}
Figure \ref{fig:pagerank} (right) compares the performance of the early fusion model as we vary the rate of fact dropout. Moderate levels of fact dropout improve performance on both datasets.
The performance increases as the fact dropout rate increases until the model is unable to learn the inference chain from KB.

\section{Conclusion}

In this paper we investigate QA using text combined with an incomplete KB, a task which has received limited attention in the past.  We introduce several benchmark problems for this task by modifying existing question-answering datasets,
and discuss two broad approaches to solving this problem---``late fusion''  and ``early fusion''. We show that early fusion approaches perform better. 

We also introduce a novel early-fusion model, called GRAFT-Net, for classifying nodes in subgraph consisting of both KB entities and text documents.  GRAFT-Net builds on recent advances in graph representation learning but includes several innovations which improve performance on this task.  
GRAFT-Nets are a single model which achieve performance competitive to state-of-the-art methods in both text-only and KB-only settings, and outperform baseline models when using text combined with an incomplete KB.
Current directions for future work include -- (1) extending GRAFT-Nets to pick spans of text as answers, rather than only entities and (2) improving the subgraph retrieval process.

\section*{Acknowledgments}
Bhuwan Dhingra is supported by NSF under grants CCF-1414030 and IIS-1250956 and by grants from Google. Ruslan Salakhutdinov is supported in part by ONR  grant N000141812861, Apple, and Nvidia NVAIL Award. 

\bibliography{emnlp2018}
\bibliographystyle{acl_natbib_nourl}

\end{document}